# A Bayesian Approach to Learning Causal Networks

**David Heckerman**
Microsoft Research, Bldg 9S/1
Redmond 98052-6399, WA
heckerma@microsoft.com

## Abstract

Whereas acausal Bayesian networks represent probabilistic independence, causal Bayesian networks represent causal relationships. In this paper, we examine Bayesian methods for learning both types of networks. Bayesian methods for learning acausal networks are fairly well developed. These methods often employ assumptions to facilitate the construction of priors, including the assumptions of parameter independence, parameter modularity, and likelihood equivalence. We show that although these assumptions also can be appropriate for learning causal networks, we need additional assumptions in order to learn causal networks. We introduce two sufficient assumptions, called *mechanism independence* and *component independence*. We show that these new assumptions, when combined with parameter independence, parameter modularity, and likelihood equivalence, allow us to apply methods for learning acausal networks to learn causal networks.

## 1 Introduction

There has been a great deal of recent interest in Bayesian methods for learning Bayesian networks from data (Spiegelhalter and Lauritzen 1990; Cooper and Herskovits, 1991, 1992; Buntine, 1991, 1994; Spiegelhalter et al. 1993; Madigan and Raftery, 1994, Heckerman et al. 1994, 1995). These methods take prior knowledge of a domain and statistical data, and construct one or more Bayesian-network models of the domain. Most of this work has concentrated on Bayesian networks interpreted as a representation of probabilistic conditional independence. Nonetheless, several researchers have proposed a causal interpretation for Bayesian networks (Pearl and Verma 1991; Spirtes et al. 1993; Heckerman and Shachter 1994). These researchers show that having a causal interpretation can be important, because it allows us to predict the affects of interventions in a domain—something that cannot be done without a causal interpretation.

In this paper, we extend Bayesian methods for learning acausal Bayesian networks to causal Bayesian networks. We offer two contributions. One, we show that acausal and causal Bayesian networks (or acausal and causal networks, for short) are significantly different in their semantics, and that it is inappropriate to blindly apply methods for learning acausal networks to causal networks. Two, despite these differences, we identify circumstances in which methods for learning acausal networks are applicable to learning causal networks.

In Section 2, we describe a causal interpretation of Bayesian networks developed by Heckerman and Shachter [1994, 1995] that is consistent with Pearl's causal-theory interpretation (e.g., Pearl and Verma [1991] and Pearl [1995a]). We show that any causal network can be represented as a special type of influence diagram. In Section 3, we review Bayesian methods for learning acausal networks, showing how various assumptions and properties—namely, parameter independence, parameter modularity, and hypothesis equivalence—facilitate the learning task. In Section 4, we show how these methods for learning acausal networks can be adapted to learn ordinary influence diagrams. In Section 5, we identify problems with this approach when learning influence diagrams that correspond to causal networks. We identify two assumptions, called *mechanism independence* and *component independence* that circumvent these problems. In Section 6, we argue that the assumption of parameter modularity is reasonable for learning causal networks, and that the property of hypothesis equivalence should be replaced with a weaker assumption called likelihood equivalence. We show that, given the assumptions of parameter independence, parameter modularity, likelihood equivalence, mechanism independence, and component independence, we can use methods for learning acausal networks to learn causal networks.

We assume that the reader is familiar the concept of random sample, the distinction between subjective and objective probability (which we call probability and physical probability, respectively), and the distinction between chance and decision variables. (We sometimes refer to a decision variable simply as a "decision.") We consider the problem of modeling relationships in a *domain* consisting of chance variables $U$ and decision variables $D$. We use lower-case letters to represent single variables and upper-case letters to represent sets of variables. We write $x = k$ to denote that variable $x$ is in state $k$. When we observe the state for every variable in set $X$, we call this set of observations a state of $X$, and write $X = k$. Sometimes, we leave the state of a variable or a set of variables implicit. We use $p(X = j|Y = k, \xi)$ to denote the (subjective) probability that $X = j$ given $Y = k$ for a person whose state of information is $\xi$; whereas, we use $pp(X = j|Y = k)$ to denote the physical probability of this conditional event.

An *influence diagram* for the domain $U \cup D$ is a model for that domain having a structural component and a probabilistic component. The *structure* of an influence diagram is a directed acyclic graph containing (square) decision and (oval) chance nodes corresponding to decision and chance variables, respectively, as well as information and relevance arcs. Information arcs, which point to decision nodes, represent what is known at the time decisions are made. Relevance arcs, which point to chance nodes, represent (by their absence) assertions of conditional independence. Associated with each chance node $x$ in an influence diagram are the probability distributions $p(x|Pa(x), \xi)$, where $Pa(x)$ are the parents of $x$ in the diagram. These distributions in combination with the assertions of conditional independence determine the joint distributions $p(U|D, \xi)$. A special kind of chance node is the deterministic node (depicted as a double oval). A node $x$ is a *deterministic node* if its corresponding variable is a deterministic function of its parents. Also, an influence diagram may contain a single distinguished node, called a *utility node* that encodes the decision maker's utility for each state of the node's parents. A utility node is a deterministic function of its predecessors and can have no children. Finally, for an influence diagram to be well formed, its decisions must be totally ordered by the influence-diagram structure. (For more details, see Howard [1981].)

An *acausal Bayesian network* is an influence diagram that contains no decision nodes (and, therefore, no information arcs). That is, an acausal Bayesian network represents only assertions of conditional independence. (For more details, see Pearl [1988].)

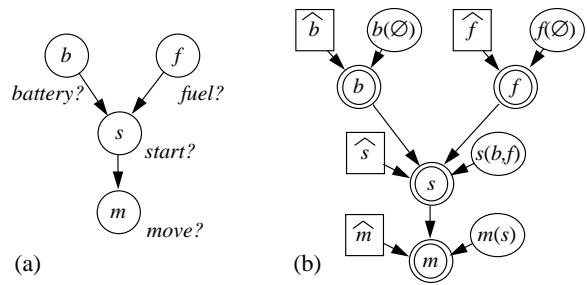

Figure 1: (a) A causal network. (b) A corresponding influence diagram. Double ovals denote deterministic nodes.

## 2 Causal Networks

In this section, we describe causal Bayesian networks and how we can represent them as influence diagrams. The influence-diagram representation that we describe is identical to Pearl's causal theory, with one exception to be discussed. Rather than present the representation directly, we follow the approach of Heckerman and Shachter (1994 and this proceedings) who define cause and effect, and then develop from this definition the influence-diagram representation of causal networks.

Roughly speaking, a causal network for a domain of chance variables $U$ is a directed acyclic graph where nodes correspond to the chance variables in $U$ and each nonroot node is the direct causal effect of its parents (Pearl and Verma, 1991). An example of a causal network is shown in Figure 1a. The diagram indicates that whether or not a car starts is caused by the condition of its battery and fuel supply, that whether or not a car moves is caused by whether or not it starts, and that (in this model) the condition of the battery and the fuel supply have no causes. In this example, we assume that all variables are binary.

Before we develop the influence-diagram representation of a causal network, we need to introduce the concepts of unresponsiveness, set decision, mapping variable, cause, causal mechanism, and canonical form. To understand the notion of unresponsiveness, consider the simple decision $d$ of whether or not to bet heads or tails on the outcome of a coin flip $c$. Let the variable $w$ represent whether or not we win. Thus, $w$ is a deterministic function of $d$ and $c$: we win if and only if the outcome of the coin matches our bet. Let us assume that the coin is fair (i.e., $p(heads|\xi) = 1/2$), and that the person who flips the coin does not know how we bet.

In this example, we are uncertain whether or not the coin will come up heads, but we are certain that whatever the outcome, it will be the same even if we choose to bet differently. We say that *c is unresponsive to d*. We cannot make the same claim about the relationship between $d$ and $w$. Namely, we know that $w$ depends on $d$ in the sense that if we bet differently then $w$ will be different. For example, we know that if we will win by betting heads then we will loose by betting tails. We say that $w$ *is responsive to* $d$.

In general, to determine whether or not chance variable $x$ is unresponsive to decision $d$, we have to answer the query "Will the outcome of $x$ be the same no matter how we choose $d$?" Queries of this form are a simple type of *counterfactual query*, discussed in the philosophical literature (e.g., Lewis [1979]). It is interesting that, in many cases, it is easy to answer such a query, even though we are uncertain about the outcome of $x$. Note that when $x$ is unresponsive to $d$, $x$ and $d$ must be probabilistically independent; whereas, the converse does not hold.

To understand the concept of a set decision, consider the chance variable *battery?* in our automobile example. Let us assume that it has only two states: "good" and "bad." Although *battery?* is a chance variable, we can imagine taking an action that will force the variable into one of its possible states. If this action has no side effects on the other variables in the model other than those required by the causal interactions in the domain, we say that we are *setting the variable*. For example, we can force the battery to fail by blowing up the car. This action, however, will also force the variable *fuel?* to become empty, and therefore does not qualify as a setting of *battery?*. In contrast, if we force the battery to fail by emptying the battery fluid on the ground, the only side effects will be those that follow from the causal interactions in the domain. Consequently, this action qualifies as a setting of *battery?*. We can extend the idea of setting a variable to a decision variable. Namely, we set a decision variable simply by choosing one of its alternatives.

A *set decision* for chance variable $x$, denoted $\hat{x}$, is a decision variable whose alternatives are "set $x$ to $k$" for each state $k$ of $x$ and "do nothing." In our example, the set decision corresponding *battery?* has three alternatives: "set the battery to be good," "set the battery to be bad," and "do nothing." Pearl and Verma (1991) introduce the concepts of setting a variable and set decision as primitives. Heckerman and Shachter (in this proceedings) formalize these concepts in terms of unresponsiveness.

To understand the concept of a mapping variable, suppose we have a collection of variables $Y$ (which may include both chance and decision variables) and a chance variable $x$. We can imagine setting $Y$ to each of its states and observing $x$—that is, observing how $Y$ maps to $x$. A *mapping variable* $x(Y)$ is a chance variable whose states correspond to all the possible mappings from $Y$ to $x$. For example, consider the variables $s$ (*start?*) and $m$ (*move?*) in our automobile example. The states of the mapping variable $m(s)$ are shown in

Table 1: The four states of the mapping variable $m(s)$.

|       | state 1 |     | state 2 |     | state 3 |     | state 4 |     |
|-------|---------|-----|---------|-----|---------|-----|---------|-----|
| start | no      | yes | no      | yes | no      | yes | no      | yes |
| move  | no      | yes | no      | no  | yes     | yes | yes     | no  |

Table 1. The first state represents the normal situation. That is, if we make the car start (in the sense of a set action), then it would move; and if we prevent the car from starting, then it would not move. The second state represents the situation where, regardless of whether or not we make the car start, the car not will move. This state would occur, for example, if a parking attendant placed a restraint on one of the car's tires. Note that, by definition, $x$ will always be a deterministic function of the mapping variable $x(Y)$ and the variables $Y$. For example, if $m(s) =$ "state 4" and $s =$ "yes," then $m =$ "no".

We can observe the mapping variable $m(s)$ directly. Namely, we can see if the car moves before and after we start the car. In general, however, mapping variables cannot be fully observed. For example, consider the decision $x$ of whether to continue or quit smoking and the chance variable $y$ representing whether or not we get lung cancer before we reach sixty years of age. In this case, we cannot fully observe the mapping variable $y(x)$, because we cannot observe whether or not we get lung cancer given both possible choices. In general, a mapping variable represents a counterfactual set of possible outcomes, only one of which we can actually observe. Rubin (1978) and Howard (1990) define concepts similar to the mapping variable.

Given these concepts, Heckerman and Shachter (1994 and this proceedings) say that *a set of variables $C$ are causes for $x$ with respect to decisions $D$ if (1) $x \notin C$ and (2) $C$ is a minimal set of variables such that $x(C)$ is unresponsive to $D$.* Roughly speaking, $C$ is a cause for $x$ with respect to $D$ if the way $C$ affects $x$ is not affected by $D$. This explication of cause is unusual in that it is conditioned on a set of decisions. Heckerman and Shachter discuss the advantages of this approach. When $C$ are causes of $x$ with respect to $D$, we call the mapping variable $x(C)$ a *causal mechanism* or simply a *mechanism*.

Given chance variables $U$ and decisions $D$, Heckerman and Shachter show that we can construct an influence diagram that represents causes for each caused variable in $U$ as follows. First, we add a node to the diagram corresponding to each variable in $U \cup D$. Next, we order the variables $x_1, \ldots, x_n$ in $U$ so that the variables unresponsive to $D$ come first. Then, for each variable $x_i$ in $U$ in order, if $x_i$ is responsive to $D$ we (1) add a causal-mechanism node $x_i(C_i)$ to the diagram, where $C_i \subseteq D \cup \{x_1, \ldots, x_{i-1}\}$, and (2) make $x_i$ a deterministic function of $C_i \cup x_i(C_i)$. Finally, we assess dependencies among the variables that are unresponsive $D$.

They show that the resulting influence diagram has the following two properties: (1) all chance nodes that are responsive to $D$ are descendants of decision nodes and (2) all nodes that are descendants of decision nodes are deterministic nodes. Influence diagrams that satisfy these conditions are said to be in *canonical form.* We note that information arcs and a utility node may be added to canonical form influence diagrams, but these constructs are not needed for the representation of cause and are not used in this discussion.

We can use an influence diagram in canonical form to represent the causal relationships depicted in a causal network. Suppose we have a set of chance variables $U$, a corresponding collection of set decisions $\hat{U}$ for $U$, and a causal network for $U$. Let $Pa(x)$ be the parents of $x$ in the causal network. Then, we can interpret the causal network to mean that, for all $x$, $Pa(x) \cup \{\hat{x}\}$ is a set of causes for $x$ with respect $\hat{U}$. Now, if we construct an influence diagram in canonical form as we have described, using an ordering consistent with the causal network, then we obtain an influence diagram where each variable $x$ is a deterministic function of the set decision $\hat{x}$, $Pa(x)$, and the causal mechanism $x(Pa(x), \hat{x})$. By the definition of a set decision, we can simplify the deterministic relationship by replacing the causal mechanism $x(Pa(x), \hat{x})$ with $x(Pa(x))$, which denotes the mappings from $Pa(x)$ to $x$ when $\hat{x}$ is set to "do nothing." For example, in our automobile domain, if $m(s) =$ state 4, $\hat{s} =$ "do nothing," and $s = yes$, then $m = no$.

The transformation from causal network to canonical form influence diagram for our automobile domain is illustrated in Figure 1. We call the variables in the original causal network *domain variables.* Each domain variable appears in the influence diagram, and is a function of its set decision $\hat{x}$, its parents in the causal network $Pa(x)$, and the mapping variable $x(Pa(x))$. (Note that $x(\emptyset) = x$ when $\hat{x} =$ "do nothing".) The mechanisms and set decisions are independent, because, as is required by canonical form, the mechanisms are unresponsive to the set decisions. Although not required by the canonical-form representation, the mechanisms are mutually independent in this example.

In general, this influence-diagram representation of a causal network is identical to Pearl's causal theory, with the exception that Pearl requires the mechanisms (which he calls *disturbances*) to be independent. One desirable consequence of this restriction is that the variables in the causal network will exhibit the conditional independencies that we would obtain by interpreting the causal network as an acausal network (Spirtes et al., 1993; Pearl, 1995a). For example, the independence of causal mechanisms in our example yield the following conditional independencies:

$$p(f|b,\xi) = p(f|\xi) \quad p(m|b,f,s,\xi) = p(m|s,\xi)$$

We obtain these same independencies when we interpret the causal network in Figure 1a as an acausal network. Nonetheless, as we shall illustrate, dependent mechanisms cannot be excluded in general.

## 3  Learning Acausal Networks

Given the correspondence in the previous section, we see that learning causal networks is a special case of learning influence diagrams in canonical form. In this section, we review methods for learning acausal Bayesian networks, such as those described by Spiegelhalter and Lauritzen (1990), Cooper and Herskovits (1991, 1992), Buntine (1991, 1994), Spiegelhalter et al., (1993), Madigan and Raftery (1994), and Heckerman et al. (1994, 1995). In the following sections, we show how these methods can be extended to learn arbitrary influence diagrams and influence diagrams in canonical form.

Suppose we have a domain consisting of chance variables $U = \{x_1, \ldots, x_n\}$. Also, suppose we have a database of cases $\mathcal{C} = \{C_1, \ldots, C_m\}$ where each case $C_l$ contains observations of one or more variables in $U$. The basic assumption underlying the Bayesian approach is that the database $\mathcal{C}$ is a random sample from $U$ with joint physical probability distribution $pp(U)$. As is done traditionally, we can characterize this physical probability distribution by a finite set of parameters $\Theta_U$. For example, if $U$ contains only continuous variables, $pp(U)$ may be a multivariate-Gaussian distribution with parameters specifying the distribution's means and covariances. In this paper, we limit our discussion to domains containing only discrete variables. Therefore, the parameters $\Theta_U$ correspond exactly to the physical probabilities in the distribution $pp(U)$. (We shall use the $\Theta$ and $pp$ notation interchangeably.)

In the general Bayesian approach to learning about these uncertain parameters, we assess prior distributions for them, and then compute their posterior distributions given the database. In the paradigm of learning acausal Bayesian networks, we add one twist to this general approach: we assume that the physical probability distribution $pp(U)$ is constrained such that it can be encoded in some acausal-network structure whose identity is possibly uncertain.

To start with a special case, let us suppose that $pp(U)$ can be encoded in some known acausal-network structure $B_s$, and that we are uncertain only about the values of the probabilities associated with this network structure. We say that *the database is a random sample from $B_s$.* Given this situation, it turns out the database $\mathcal{C}$ can be separated into a set of random samples, where these random samples are determined by the structure of $B_s$. For example, consider the domain consisting of two variables $x$, where each variables has possible states 0 and 1. Then, the asser-tion that the database is a random sample from the structure $x \to y$ is equivalent to the assertion that the database can be separated into at most three random samples: (1) the observations of $x$ are a binomial sample with parameter $\theta_{x=1}$, (2) the observations of $y$ in those cases (if any) where $x = 0$ are a binomial sample with parameter $\theta_{y=1|x=0}$, and (3) the observations of $y$ in those cases (if any) where $x = 1$ are a binomial sample with parameter $\theta_{y=1|x=1}$. Figure 2a contains an acausal network that illustrates some of the conditional independencies among the database cases and network parameters for this assertion.

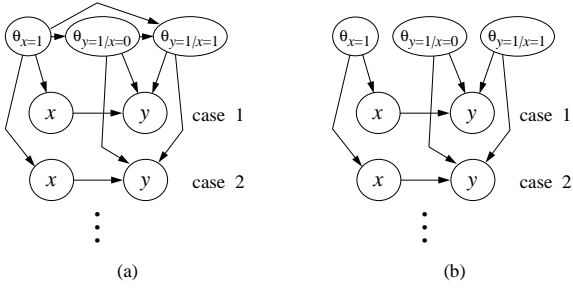

Figure 2: (a) Conditional independencies associated with the assertion that the database is a random sample from the structure $x \to y$, where $x$ and $y$ are binary. (b) The additional assumption of parameter independence.

Given this decomposition into random samples, we can update each parameter independently under two conditions: (1) the parameters are independent, an assumption we call *parameter independence,* and (2) the database is *complete* (i.e., every variable is observed in every case). The assumption of parameter independence is illustrated in Figure 2b.

Let us examine this updating for an arbitrary acausal-network structure $B_s$ for domain $U$. We discuss the situation where data may be missing later in this section. Let $r_i$ be the number of states of variable $x_i$; and let $q_i = \prod_{x_l \in Pa(x_i)} r_l$ be the number of states of $Pa(x_i)$. Let $\theta_{ijk}$ denote the parameter corresponding to the physical probability $p(x_i = k|Pa(x_i) = j, \xi)$ ($\theta_{ijk} > 0$; $\sum_{k=1}^{r_i} \theta_{ijk} = 1$). In addition, we define

$$\Theta_{ij} \equiv \cup_{k=1}^{r_i} \{\theta_{ijk}\} \qquad \Theta_{Bs} \equiv \cup_{i=1}^{n} \cup_{j=1}^{q_i} \Theta_{ij}$$

That is, the parameters $\Theta_{Bs}$ correspond to the physical probabilities of the acausal-network structure $B_s$.

To illustrate the updating approach, suppose that each variable set $\Theta_{ij}$ has a Dirichlet distribution:

$$p(\Theta_{ij}|B_s^h, \xi) = c \cdot \prod_{k=1}^{r_i} \theta_{ijk}^{N'_{ijk}-1} \qquad (1)$$

where $B_s^h$ is the assertion (or "hypothesis") that the database is a random sample from the network structure $B_s$, and $c$ is some normalization constant. Then, given parameter independence and a complete database, if $N_{ijk}$ is the number of cases in database $\mathcal{C}$ in which $x_i = k$ and $Pa(x_i) = j$, we obtain

$$p(\Theta_{ij}|\mathcal{C}, B_s^h, \xi) = c \cdot \prod_k \theta_{ijk}^{N'_{ijk}+N_{ijk}-1} \qquad (2)$$

where $c$ is some other normalization constant. Furthermore, taking the expectation of $\theta_{ijk}$ with respect to the distribution for $\Theta_{ij}$ for every $i$ and $j$, we obtain the probability that each $x_i = k$ and $Pa(x_i) = j$ in $C_{m+1}$ (the next case $C_{m+1}$ to be seen after seeing the database):

$$p(C_{m+1}|\mathcal{C}, B_s^h, \xi) = \prod_{i=1}^{n} \prod_{j=1}^{q_i} \frac{N'_{ijk} + N_{ijk}}{N'_{ij} + N_{ij}} \qquad (3)$$

where $N'_{ij} = \sum_{k=1}^{r_i} N'_{ijk}$ and $N_{ij} = \sum_{k=1}^{r_i} N_{ijk}$.

Now, suppose we are not only uncertain about the probabilities, but also uncertain about the structure that encodes them. We express this uncertainty by assigning a prior probability $p(B_s^h|\xi)$ to each possible hypothesis $B_s^h$, and update these probabilities as we see cases. In so doing, we learn about the structure of the domain. From Bayes' theorem, we have

$$p(B_s^h|\mathcal{C}, \xi) = c \; p(B_s^h|\xi) \; p(\mathcal{C}|B_s^h, \xi) \qquad (4)$$

where $c$ is a normalization constant. Also, from the product rule, we have

$$p(\mathcal{C}|B_s^h, \xi) = \prod_{l=1}^{m} p(C_l|C_1, \ldots, C_{l-1}, B_s^h, \xi) \qquad (5)$$

We can evaluate each term on the right-hand-side of this equation using Equation 3, under the assumption that the database $\mathcal{C}$ is complete. For the posterior probability of $B_s^h$ given $\mathcal{C}$, we obtain

$$p(B_s^h|\mathcal{C}, \xi) = c \cdot p(B_s^h|\xi) \qquad (6)$$
$$\cdot \prod_{i=1}^{n} \prod_{j=1}^{q_i} \frac{\Gamma(N'_{ij})}{\Gamma(N'_{ij}+N_{ij})} \cdot \prod_{k=1}^{r_i} \frac{\Gamma(N'_{ijk}+N_{ijk})}{\Gamma(N'_{ijk})}$$

Using these posterior probabilities and Equation 3, we can compute the probability distribution for the next case to be observed after we have seen a database. From the expansion rule, we obtain

$$p(C_{m+1}|\mathcal{C}, \xi) = \sum_{B_s^h} p(C_{m+1}|\mathcal{C}, B_s^h, \xi) \; p(B_s^h|\mathcal{C}, \xi) \qquad (7)$$

When the database contains missing data, we can compute $p(B_s^h|\mathcal{C}, \xi)$ exactly, by summing the result of Equation 5 over all possible completions of the database (see Section 7). Unfortunately, this approach is intractable when many observations are missing. Consequently, we often use approximate methods such as filling in missing data based on the data that is present (Titterington, 1976; Cowell et al., 1995), the EM algorithm (Dempster et al., 1977), and Gibbs sampling (York, 1992; Madigan and Raftery, 1994).

When we believe that only a few network structures are possible, the approach we have discussed is essentially all there is to learning network structure. Namely, we directly assess the priors for the possible network structures and their parameters, and subsequently use Equations 3 and 7 or their generalizations for continuous variables and missing data. Nonetheless, the number of network structures for a domain containing $n$ variables is more than exponential in $n$. Consequently, when we cannot exclude almost all of these network structures, we need efficient methods for assigning priors to structures and parameters (e.g., Buntine [1991], Spiegelhalter et al. [1993], and Heckerman et al. [1995]), as well as search methods for identifying structures that contribute significantly to the sum in Equation 3 (e.g., Cooper and Herskovits [1992] and Heckerman et al. [1995]).

Here, we review an efficient method described by Heckerman et al. [1995] for assigning priors to the parameters of all possible network structures. In their approach, a user assesses a *prior network*: an acausal Bayesian network for the first case to be seen in database, under the assumption that there are no constraints on the parameters. More formally, this prior network represents the joint probability distribution $p(C_1|B_{sc}^h, \xi)$, where $B_{sc}$ is any network structure containing no missing arcs. Then, the user assesses an *equivalent sample size* $N'$ for this prior network. ($N'$ is a measure of the user's confidence in his assessment of the prior network.) Then, for any given network structure $B_s$, where $x_i$ has parents $Pa(x_i)$, we compute the Dirichlet exponents in Equation 1 using the relation

$$N'_{ijk} = N' \cdot p(x_i = k, Pa(x_i) = j|B_{sc}^h, \xi) \qquad (8)$$

where the probability is computed from the prior network.

Heckerman et al. [1995] derive this approach from the assumption of parameter independence, an additional assumption called *parameter modularity*, and a property called *hypothesis equivalence*. The property of hypothesis equivalence stems from the fact that two acausal-network structures can be *equivalent*—that is, represent exactly the same sets of probability distributions (Verma and Pearl, 1990). For example, for the three variable domain $\{x, y, z\}$, each of the network structures $x \to y \to z$, $x \leftarrow y \to z$, and $x \leftarrow y \leftarrow z$ represents the distributions where $x$ and $z$ are conditionally independent of $y$, and are therefore equivalent. Given the definition of the hypothesis $B_s^h$, it follows that the hypotheses corresponding to two equivalent structures must be the same, which is the property of hypothesis equivalence.

The assumption of parameter modularity says that, given two network structures $B_{s1}$ and $B_{s2}$, if $x_i$ has the same parents in $B_{s1}$ and $B_{s2}$, then

$$p(\Theta_{ij}|B_{s1}^h, \xi) = p(\Theta_{ij}|B_{s2}^h, \xi)$$

for $j = 1, \ldots, q_i$. Heckerman et al. [1995] call this property parameter modularity, because it says that the distributions for parameters $\Theta_{ij}$ depend only on the structure of the network that is local to variable $x_i$—namely, $\Theta_{ij}$ only depends on $x_i$ and its parents. In Section 6, we examine the appropriateness of hypothesis equivalence and parameter modularity for learning causal networks.

## 4 Learning Influence Diagrams

Before we consider the problem of learning influence diagrams that correspond to causal networks, let us examine the task of learning arbitrary influence diagrams.

This task is straightforward once we make the following observations. One, by the definitions of information arc and utility node, information arcs and the predecessors of a utility node are known with certainty by the decision maker and, therefore, are not learned.[1] Thus, we need only learn the relevance-arc structure and the physical probabilities associated with chance nodes. Two, by definition of a decision, the states of all decision variables are known by the decision maker in every case. Thus, assuming these decisions are recorded, we have complete data for $D$ in every case of the database.

Given these observations, it follows that the problem of learning influence diagrams for the domain $U \cup D$ reduces to the problem of learning acausal Bayesian networks for $U \cup D$, where we interpret the decision variables $D$ as chance variables. The only caveat is that the learned relevance-arc structures will be constrained by the influence-diagram semantics. In particular, a relevance-arc structure is eligible to be learned (i.e., has a corresponding hypothesis that can have a non-zero prior) if and only if (1) every node in $D$ is a root node and (2) that structure when combined with the information-arc structure declared by the decision maker contains no directed cycles. (Note that both of these constraints are satisfied by canonical-form representations of causal networks.)

---

[1] For simplicity of presentation, we assume that information-arc and utility-node structure is identical for all cases in the database.

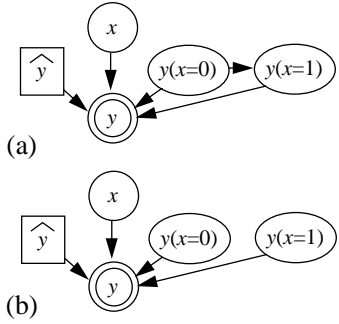

Figure 3: (a) A decomposition of the mapping variable $y(x)$. (b) The assumption of component independence.

## 5 Learning Causal-Network Parameters

In this section, we consider aspects of learning influence diagrams peculiar to influence diagrams in canonical form. In this discussion, we assume that the structure of the influence diagram is known, and that we need to learn only the parameters of the structure.

One difficulty associated with learning influence diagrams in canonical form occurs in domains where we can set variables only once (or a small number of times) so that the mechanisms are not fully observable. For example, recall our decision to continue or quit smoking where $x$ denotes our decision and $y$ denotes whether or not we get lung cancer before the age of sixty. In this case, we cannot fully observe the mapping variable $y(x)$, because we cannot observe whether or not we get lung cancer for both possible choices. Given any one choice for $x$ and observation of $y$, we exclude only two of the four states of $y(x)$. Consequently, it would seem that learning about $y(x)$ would be difficult if not impossible.

We can understand this difficulty in another way. Given any mapping variable $y(X)$ where $X$ has $q$ states, we can decompose $y(X)$ into a set of variables $y(X = k_1), \ldots, y(X = k_q)$, where variable $y(X = k)$ represents the variable $y$ when $X$ is set to state $k$. We call these variables *mechanism components*. For example, Figure 3a illustrates the components of the mechanism variable $y(x)$, where $x$ is a binary variable. Note that, by the definition of a mechanism component, we have

$$pp(y(X = k)) = pp(y|X = k) \qquad (9)$$

An analogous equation holds for (subjective) probabilities.

Given this decomposition, the a setting of $X$ and the observation of $y$ is equivalent to the observation of exactly one of the components of $y(X)$. Thus, if we can set $X$ only once, as in the smoking example, we cannot observe multiple mechanism components. Consequently, we cannot learn about the physical probabilities that characterize the dependencies among the components. Holland (1986) calls this problem, albeit in a different mathematical formalism, the "fundamental problem with causal inference."

To circumvent this problem, we can assume that mechanism components are independent, an assumption we call *component independence*.[2] If this assumption is incorrect, then we will not learn correct counterfactual relationships. Regardless of the assumption's correctness, however, we can correctly quantify the affects of a single setting action.

For example, in the smoking decision, the mechanism components are clearly dependent: Knowing that we quit and got lung cancer ($y(x = 0) = 1$) makes it more likely that we would have gotten lung cancer had we continued ($y(x = 1) = 1$). Nonetheless, suppose we assume the components are independent and learn the physical probabilities from a database of cases. Then, although we learn incorrect counterfactual relationships—namely, that $y(x = 0)$ and $y(x = 1)$ are independent—we can still learn the correct marginal physical probabilities associated with both mechanism components. Thus, by Equation 9, we can learn the correct physical probability that that we will get cancer if we continue to smoke as well as the correct physical probability of cancer if we quit smoking.

A second complication with learning influence diagrams in canonical form is the possible dependency among different mechanisms. For example, suppose we model the voltages in a logic circuit containing two buffers in series as shown in Figure 4a. Here, $x$ and $z$ represent the input and output voltages of the circuit, respectively, and $y$ represents the voltage between the two buffers. The causal network for this circuit is $x \rightarrow y \rightarrow z$. The corresponding influence diagram in canonical form is shown in Figure 4b. The causal mechanism $y(x)$ represents the possible mappings from the input to the output of the first buffer. The possible states of $y(x)$ are "output normal," "output always zero," "output always one," and "output inverted". That is, this causal mechanism is a representation of the working status of the buffer. Similarly, the mapping variable $z(y)$ represents the working status of the second buffer. Thus, these mechanisms will be dependent whenever buffer function is dependent—for example, when it is possible for the circuit to overheat and cause both buffers to fail.

---

[2]We note that, from Equation 9, under the assumption of component independence, we can fill in the probability tables associated with the canonical-form representation of a causal network by copying the probabilities associated with that causal network. Without this assumption, the canonical-form representation requires additional probability assessments.

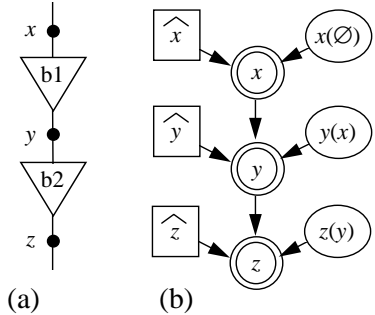

Figure 4: (a) A logic circuit containing two buffers in series. (b) A causal network for the circuit, represented as an influence diagram in canonical form.

Dependent mechanisms lead to practical problems. Namely, given the large number of states typically associated with mapping variables, the assessment of priors is difficult, and we require vast amounts of data to learn. Fortunately, we can often introduce additional domain variables in order to render mechanisms independent. In our circuit example, if we add to our domain the variable $t$ representing the temperature of the circuit, then the new mechanisms $y(x,t)$ and $z(y,t)$ will be independent. This solution sometimes creates a another problem with learning: we may not be able to observe the variables we introduce. We address this issue in Section 7.

Given mechanism independence and component independence for all mechanisms, the only chance variables that remain in a canonical form influence diagram are mutually independent mechanism components. Consequently, if we also assume parameter independence, then the problem of learning a causal network essentially reduces that of learning an acausal network.

To illustrate this equivalence, consider again our two binary-variable domain, and assume that the database is a random sample from an influence diagram corresponding to the causal network $x \to y$. Given the assumptions of mechanism, component, and parameter independence, we have the influence diagram in Figure 5a, where the deterministic functions for $x$ and $y$ are given by

$$x = \begin{cases} x(\emptyset) & \text{if } \hat{x} = \text{``do nothing''} \\ k & \text{if } \hat{x} = \text{``set x to } k\text{''} \end{cases}$$

$$y = \begin{cases} y(x=j) & \text{if } \hat{y} = \text{``do nothing'' and } x = j \\ k & \text{if } \hat{y} = \text{``set y to } k\text{''} \end{cases}$$

by the definitions of set decision and mechanism component.

Now, suppose that all the set decisions are "do nothing." In this situation, if we integrate out the mechanism variables from the diagram (as discussed in Shachter [1986]), then we obtain the influence diagram shown in Figure 5b. This structure is equivalent to the one shown in Figure 2b for learning the acausal network $x \to y$. Thus, we can update the parameters of the causal network $x \to y$ just as update those for the corresponding acausal network.

This result generalizes to arbitrary causal networks. In particular, if all set decisions in a particular case $C_l$ are "do nothing," we say the that observations of the domain variables in $C_l$ are *non-experimental data*. Otherwise, we say that the observations are *experimental data*. Given a case of non-experimental data, we update the parameters of a causal network just as we would the parameters of the corresponding acausal network (assuming mechanism, component, and parameter independence).

The updating procedure for experimental data is slightly different from that for non-experimental data. In our two-variable example, if we set $y$ and observe $x$ (with $\hat{x}$ set to "do nothing"), then we obatin the influence diagram shown in Figure 5c. Here, the arcs to $y$ are removed, because we have set the variable $y$. Consequently, neither $\theta_{y=1|x=0}$ nor $\theta_{y=1|x=1}$ are updated given this data. In general, to update the parameters for a canonical form influence diagram given experimental data where we have set $x_i$, we break all arcs to $x_i$, and update the parameters as we would for an acausal network.

## 6 Learning Causal-Network Structure

In Section 3, we saw that, given the assumptions of parameter independence, parameter modularity, and hypothesis equivalence, we can assess priors for the parameters of all possible acausal-network structures by constructing a single prior network for the first case to be seen in the database and assessing an equivalent sample size (confidence) for this prior network. Thus, given the discussion in the previous section, it follows that we can use this prior-network methodology to establish priors for causal-network learning, provided we assume mechanism independence, component independence, parameter independence, parameter modularity, and hypothesis equivalence. In this section, we examine the assumptions of parameter modularity and likelihood equivalence for learning causal networks.

The assumption of parameter modularity has a compelling justification in the context of causal networks. Namely, suppose a domain variable $x$ has the same parents $Pa(x)$ in two possible causal-network structures. Then, it is reasonable to believe that the causal mechanism $x(Pa(x))$ should be the same given either structure. It follows that its parameters $\Theta_{x|Pa(x)}$ for both structures must have the same prior distributions— that is, parameter modularity must hold.

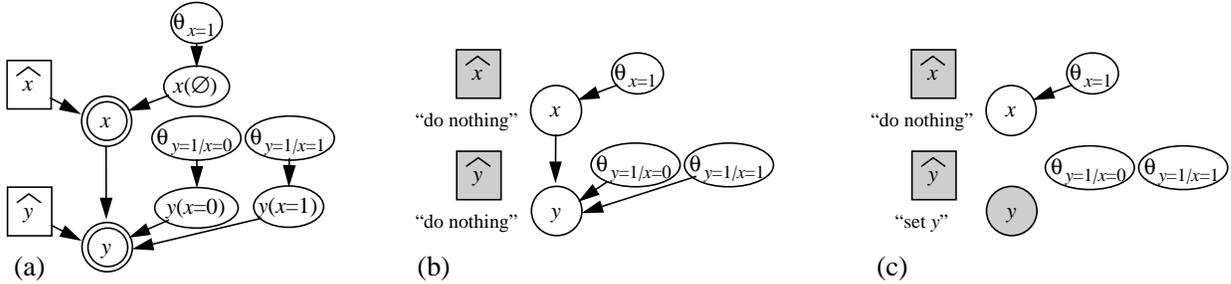

Figure 5: (a) Mechanism independence, component independence, and parameter independence associated with the causal network $x \to y$. (b,c) Corresponding diagrams when (b) $\hat{x}$ and $\hat{y}$ are "do nothing" and (c) $\hat{x} =$"do nothing" and $\hat{y} \neq$"do nothing."

In contrast, the property of hypothesis equivalence cannot be applied to causal networks. For example, in our two-variable domain, the causal network $x \to y$ represents the assertion that $x$ causes $y$, whereas the causal network $y \to x$ represents the assertion that $y$ causes $x$. Now, it is possible for both $x$ to cause $y$ and vice versa when the two variables are somehow deterministically related (e.g., consider the variables pressure and volume in a closed physical system). Barring such deterministic relationships, however, the hypotheses corresponding to these two network structures are mutually exclusive. Consequently, hypothesis equivalence does not hold.

Nonetheless, when we know little about the structure of a domain, we have often found it reasonable to assume that data cannot help to distinguish between equivalence network structures. To express this assumption formally, let $\Theta_U$ denote the parameters of the joint space, and let $C_s^h$ denote the hypothesis that the database is a random sample from the influence diagram corresponding to the causal-network structure $C_s$. Then, we have

$$p(\Theta_U | C_{s1}^h, \xi) = p(\Theta_U | C_{s2}^h, \xi)$$

whenever the causal-network structures $C_{s1}$ and $C_{s2}$ are equivalent (when interpreted as acausal networks). We call this assumption *likelihood equivalence*. Heckerman et al. [1995] show that the prior-network methodology is still justified when we replace the assumption of hypothesis equivalence with that of likelihood equivalence.

Under the assumptions of mechanism, component, and parameter independence, the assumption of likelihood equivalence has an interesting characterization. Consider again our two-variable domain. Suppose we know nothing about the domain, having uninformative Dirichlet priors on the parameters of both network structures (all Dirichlet exponents arbitrarily close to $-1$). Further, suppose we adopt the assumption of likelihood equivalence for the two network structures $x \to y$ and $y \to x$. Now, suppose we obtain a single case of experimental data where we set $x = 1$ and observe $y = 1$. According to our updating procedure described in the previous section, for the network structure $x \to y$, we update the parameter $\theta_{y=1|x=1}$, but not the parameter $\theta_{x=1}$. In contrast, for the network structure $y \to x$, we update the parameter $\theta_{y=1}$, but not the parameter $\theta_{x=1|y=1}$. As a result, our posterior distributions for $\Theta_U = \{\theta_{x=0,y=0}, \theta_{x=0,y=1}, \theta_{x=1,y=0}, \theta_{x=1,y=1}\}$ will no longer satisfy likelihood equivalence. One can show that, for any domain, if we have an uninformative Dirichlet prior for that domain and we are given a database containing experimental data, then the resulting posterior distributions for $\Theta_U$ will violate likelihood equivalence. Therefore, we can assess whether or not likelihood equivalence holds by asking ourselves whether or not our prior knowledge is equivalent to having seen only non-experimental data.

We note that the assumption of likelihood equivalence tends to be less reasonable for more familiar domains. For example, a doctor may be uncertain as to whether disease $d_1$ causes disease $d_2$ or vice versa, but he may have well-defined hypotheses about why $d_1$ causes disease $d_2$ and vice versa. In this case, the assumption of likelihood equivalence would likely be unreasonable.

We emphasize that experimental data can be crucial for learning causal structure. In our two-variable domain, suppose we believe that either $x$ causes $y$ or $y$ causes $x$. Then, if we set $x$ to different states and learn that the probability of $y$ depends on $x$, then we learn that $x$ causes $y$. To verify this relation, we can set $y$ to different states and check that the probability of $x$ remains the same. Conversely, if we set $y$ to different states and learn that the probability of $x$ depends on $y$, then we learn that $y$ causes $x$.

Also, we may need experimental data to quantify the effects of intervention—for example, to learn the physical probability distribution $pp(y|\hat{x} = 1)$. Given a causal structure, however, there are situations where we can quantify the effects of intervention using observational data only (Pearl and Verma, 1991; Pearl, 1995a).

# 7 Learning Hidden Variables

In Section 5, we saw that we could often remove dependencies between causal mechanisms by adding additional domain variables. In many situations, however, we can never observe these variables. We say that these variables are *hidden*.

As we have discussed, methods for learning acausal networks with missing data are known (e.g., exact, EM, Gibbs sampling). These methods can be applied to databases containing hidden variables. Thus, under the assumptions of mechanism independence, component independence, parameter independence, parameter modularity, and likelihood equivalence, we can learn causal networks with hidden variables using these methods in conjunction with the prior-network methodology.

To illustrate this approach, let us consider a simple medical domain containing two observable variables $h$ and $l$ representing the presence or absence of heart disease and lung disease, respectively, and a hidden variable $g$ representing the presence or absence of a gene that predisposes one to both diseases. Two possible causal-network structures for this domain are shown in Figure 6. In the network structure labeled $C_{s1}$, $h$ causes $l$, and $g$ is a hidden common cause of both diseases. In $C_{s2}$, the disease variables are related only through the hidden common cause. Suppose that only these two network-structure hypotheses are possible and that they are equally likely a priori. In addition, suppose our prior network for this domain is $C_{s2}$ with the probabilities shown in Figure 6, and $N'$ (the equivalent sample size for this network) is 24. Finally suppose we have a database $\mathcal{C}$ containing two cases where—in both cases—all set decisions are "do nothing," $h = 1$ (heart disease present), and $l = 1$ (lung disease present).

Because there are only two cases, we can compute the posterior probabilities of both network-structure hypotheses exactly, using Equation 7 (which applies to complete databases), Equation 8, and the relation

$p(\mathcal{C}|C_s^h,\xi) =$
$\quad p(g_1=1, h_1=1, l_1=1, g_2=1, h_2=1, l_2=1|C_s^h,\xi) +$
$\quad p(g_1=0, h_1=1, l_1=1, g_2=1, h_2=1, l_2=1|C_s^h,\xi) +$
$\quad p(g_1=1, h_1=1, l_1=1, g_2=0, h_2=1, l_2=1|C_s^h,\xi) +$
$\quad p(g_1=0, h_1=1, l_1=1, g_2=0, h_2=1, l_2=1|C_s^h,\xi)$

where the subscripts on the variables denote case numbers. For example, from Equations 7 and 8, the first term in this sum for $C_{s1}$ is given by

$$\frac{\Gamma(24)}{\Gamma(26)}\frac{\Gamma(14)}{\Gamma(12)}\frac{\Gamma(12)}{\Gamma(14)}\frac{\Gamma(8)}{\Gamma(6)}\frac{\Gamma(6)}{\Gamma(8)}\frac{\Gamma(5)}{\Gamma(3)} = 1/50$$

Performing the sums and applying Bayes' theorem, we obtain $p(C_{s1}^h|\mathcal{C},\xi) = 0.51$ and $p(C_{s2}^h|\mathcal{C},\xi) = 0.49$.

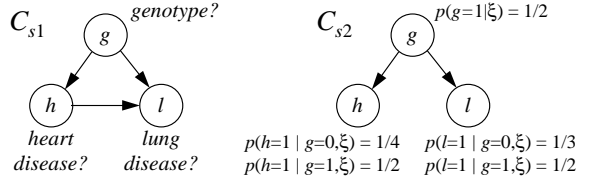

Figure 6: Two possible causal networks that explain an observed dependence between heart and lung disease.

For domains containing hidden variables, Pearl (1995b) has suggested a generalization of the assumption of likelihood equivalence, which says that if two causal networks are equivalent with respect to the distributions they encode for the *observed* variables, then the parameters for those observed variables should have identical priors. We call this property *strong likelihood equivalence*. This property does not hold in our simple medical example. Namely, the two network structures $C_{s1}$ and $C_{s2}$ are equivalent with respect to the variables $h$ and $l$ (i.e., both structures can represent any joint distribution over these variables). Nonetheless, as we saw in the previous example, observations can help to discriminate the two network structures. Thus, given the assumptions of mechanism and component independence, strong likelihood equivalence is not consistent with our prior-network methodology. That is, strong likelihood equivalence is not consistent with the assumptions of parameter independence and parameter modularity. Consequently, strong likelihood equivalence may lead to a method for assessing priors on parameters that is an alternative to the prior-network approach.

# 8 Learning More General Causal Models

Our presentation has concentrated on domains where all variables (except root nodes) have causes. We emphasize that this restriction is unnecessary, given the definition of cause given by Heckerman and Shachter (1994 and this proceedings). In particular, as shown by these researchers, the relationships in domains where only some variables have causes can be encoded in canonical form. Consequently, we can often apply the learning methods we have described to these more general domains.

# Acknowledgments

This work was motivated by conversations with Max Chickering, Greg Cooper, Dan Geiger, and Judea Pearl. Jack Breese and Max Chickering provided useful suggestions on earlier versions of this manuscript.